\documentclass{article}
\usepackage{PRIMEarxiv}

\usepackage[colorlinks,urlcolor=blue,linkcolor=blue,citecolor=blue]{hyperref}

\usepackage{color,array}

\usepackage{graphicx}
\usepackage{cite}

\usepackage{xcolor, soul, tikz, svg, subcaption, adjustbox, float}
\usepackage{array}
\usepackage{caption}
\usepackage{makecell}
\usepackage{multirow}
\usepackage{tabularx}
\usepackage{amsmath}
\usepackage{tcolorbox}
\usepackage{verbatim}
\usepackage{booktabs}
\usepackage{hyperref}       % hyperlinks
\usepackage{url}            % simple URL typesetting
\usepackage{amsfonts}       % blackboard math symbols
\usepackage{nicefrac}       % compact symbols for 1/2, etc.
\usepackage{microtype}      % microtypography
\usepackage{lipsum}
\usepackage{graphicx}       % graphics
\graphicspath{{media/}}
\usepackage{colortbl}
\usepackage[table]{xcolor}
\usepackage{colortbl}
\usepackage[table,xcdraw]{xcolor} % safer and more color options

\usepackage{soul}
\sethlcolor{yellow}  % Optional: sets the highlight color
\usepackage{pifont}

\usepackage{xcolor}

\usepackage{soul}
\sethlcolor{yellow}  % Optional: sets the highlight color

\usepackage{booktabs}   % for clean tables
\usepackage{tabularx}   % for width-flexible tables

\definecolor{cust-yellow}{HTML}{e78700}
\definecolor{cust-green}{HTML}{5aab31}
\definecolor{cust-blue}{HTML}{3074ae}

\definecolor{lime}{HTML}{A6CE39}
\DeclareRobustCommand{\orcidicon}{
	\begin{tikzpicture}
	\draw[lime, fill=lime] (0,0) 
	circle [radius=0.16] 
	node[white] {{\fontfamily{qag}\selectfont \tiny ID}};
	\draw[white, fill=white] (-0.0625,0.095) 
	circle [radius=0.007];
	\end{tikzpicture}
	\hspace{-2mm}
}
\foreach \x in {A, ..., Z}{\expandafter\xdef\csname orcid\x\endcsname{\noexpand\href{https://orcid.org/\csname orcidauthor\x\endcsname}
			{\noexpand\orcidicon}}
}

\title{PersoBench: Benchmarking Personalized Response Generation in Large Language Models}

\author{
    Saleh Afzoon$^{1}$ \orcidA{}, 
    Zahra Jamali$^{2}$ \orcidD{},
    Usman Naseem$^{1}$ \orcidB{}, 
    Amin Beheshti$^{1}$ \orcidC{}\\
    $^{1}$School of Computing, Macquarie University, Sydney, Australia \\
    $^{2}$School of Electrical and Computer Engineering, Shiraz University, Shiraz, Iran \\
    \small \texttt{saleh.afzoon@hdr.mq.edu.au}, \small \texttt{zahrajamali@hafez.shirazu.ac.ir}, \small \texttt{ \{usman.naseem, amin.beheshti\}@mq.edu.au}
}

\begin{document}
\maketitle

\begin{abstract}

While large language models (LLMs) have exhibited impressive conversational capabilities, their proficiency in delivering personalized responses remains unclear. Although recent benchmarks automatically evaluate persona consistency in role-playing contexts using LLM-based judgment, the evaluation of personalization in response generation remains underexplored. To address this gap, we present an automated benchmarking pipeline, PersoBench, to evaluate the personalization ability of LLMs in persona-aware dialogue generation within a zero-shot setting. 
Our framework employs a structured pipeline comprising speaker-aware annotation, task-specific and context-driven prompt construction, response post-processing, and automated evaluation across multiple dimensions of generation quality. In particular, the pipeline performs text preprocessing and speaker labeling, constructs structured prompts with task instructions and LLM roles, validates response format, and evaluates valid outputs across fluency, personalization, diversity, and coherence. We assess the performance of four open-source and four closed-source LLMs using well-known datasets and a range of explicit metrics. Our findings reveal that while LLMs excel at generating fluent and diverse responses, they are far from satisfactory in delivering personalized and coherent responses, considering both the conversation context and the provided personas.

\end{abstract}


\section{Introduction}

Large Language Models (LLMs) have revolutionized NLP, excelling in human-like text generation across domains and becoming central to dialogue systems.
Yet, as LLMs are increasingly deployed in real-world applications, such as customer support \cite{bink2024personalized}, mental health counseling \cite{guo2024large, kang2025development}, and educational tutoring \cite{sharma2025role}, the ability to generate personalized responses is no longer optional. Personalization enhances user satisfaction, trust, and task success by adapting responses to individual preferences, goals, and conversational history \cite{loh2023harnessing}. Without personalization, even fluent and coherent outputs risk feeling generic or misaligned with user intent.
Despite this growing demand, evaluating LLMs' capacity for personalization remains underexplored. Existing benchmarks focus on role adherence or character simulation, such as RPBench-Auto \cite{rpbench}, TIMECHARA \cite{ahn2024timechara}, and RoleLLM \cite{wang2023rolellm}, but none offer a dedicated framework for assessing automatic personalized response generation. Moreover, these benchmarks often rely on LLMs as evaluators, introducing bias, and suffer from limited experimental scope \cite{ferdousi2025rhealthtwin, yuan2025science, rpbench}. This paper addresses these gaps by proposing a benchmarking pipeline specifically designed to evaluate personalized response generation, with robust evaluation protocols and diverse user contexts.

%% 7- Brief contributions
To fill this gap, we introduce \textbf{PersoBench}, an automated benchmarking pipeline for response personalization, to assess the strengths and limitations of current LLMs in generating personalized responses. To the best of our knowledge, no prior work has introduced a comprehensive pipeline specifically focused on evaluating response personalization in LLMs. Using comprehensive datasets and a diverse set of established metrics, including fluency, diversity, coherence, and personalization, we ensure a robust evaluation of various aspects of response generation, drawing on insights from a recent survey in the field \cite{chen2024recent}. While fluency and diversity assess the fundamental requirements of a dialogue system, coherence and personalization evaluate response alignment with the provided user persona.
More specifically, in line with this objective of the mentioned context, we aim to answer the following research questions:

\begin{enumerate}

  \item To what extent can LLMs generate natural responses that are linguistically fluent? 
  
  \item How proficient are LLMs at generating diverse responses?
  
  \item How effectively can LLMs generate personalized responses that reflect the provided persona?

  \item To what extent do LLMs generate coherent responses aligned with the dialogue context and persona?
  
\end{enumerate}

% Datasets and LLMs
% To answer the above questions, we benchmarked four open-source and four closed-source LLMs across three popular persona-aware conversational datasets.
To answer the above questions, we benchmarked four open-source and four closed-source LLMs across three popular persona-aware conversational datasets, each highlighting a distinct facet of personalization such as open-domain interests, structured traits, or task-specific roles. This multi-perspective design enables a more robust and generalizable evaluation of personalized response generation.
The LLMs' performance are evaluated in a zero-shot setting to benchmark their inherent capabilities without introducing examples that could enhance their performance, ensuring the results reflect their baseline ability. This approach is applied to both standard and Chain of Thought (CoT) reasoning configurations. Our evaluation goes beyond assessing overall response quality and contextualization, employing eight established metrics in conversational AI to gauge performance across fluency, diversity, coherence, and personalization dimensions.
This work contributes by (1) offering a multi-aspect evaluation of LLM personalization in zero-shot settings, (2) investigating the influence of CoT prompting on response personalization, and (3) assessing response time and instructability (the ability to adhere to task instructions) based on LLMs' adherence to task instructions.

Our findings reveal that response personalization remains a challenging task for LLMs, highlighting the need for further research to enhance their performance in this regard.

The implementation of PersoBench, including evaluation scripts and results, is publicly available at: \href{https://github.com/salehafzoon/PersoBench}{\texttt{github.com/salehafzoon/PersoBench}}.

%%%%%%%%%%%%%%%%%%%%%%%%%%%%%%%%%%%%%%%%%%%%%%%%%%%%%%%%%%%%%%%%%%%%%%%

\section{Related Work}\label{sec2}

\subsection{Persona-aware Response Generation}

Personalized response generation has emerged as a key challenge in open-domain dialogue systems, where maintaining consistency with user-specific traits is essential for engagement and realism~\cite{liu2025persona}. A range of modeling strategies has been proposed to address this challenge, focusing on persona conditioning, stylistic control, and long-term coherence.

To mitigate temporal drift in character-driven interactions, TIMECHARA~\cite{ahn2024timechara} is introduced to model time-sensitive persona attributes. By encoding temporal markers and character evolution, this approach improved consistency across multi-turn dialogues. RoleLLM~\cite{wang2023rolellm} extended this idea by incorporating RoleBench and RoCIT, which enabled role-conditioned generation through structured prompts and contextual role embeddings. Evaluation logic was embedded within the generation pipeline, allowing for internal consistency checks.
Stylistic personalization was explored in RAGs to Style~\cite{neelakanteswara2024rags}, where retrieval-augmented generation was combined with style embeddings to guide output fluency and tone. This method retrieved relevant persona-aligned examples and fused them with stylistic constraints during decoding. In contrast, PAPER~\cite{li2024persona} adopted a CoT framework to interpret and expand persona tags prior to generation, enhancing transparency and interpretability. PPlug~\cite{liu2024llms+} proposed a plug-and-play module that attaches user embeddings to LLM inputs, enabling scalable personalization without retraining or fine-tuning.

These approaches reflect a spectrum of personalization strategies, including prompt engineering, retrieval-based conditioning, modular embedding, and reasoning over persona traits. Despite their innovation, evaluation remains fragmented. Metrics such as Persona-F1 \cite{jiang2020pednet}, AlignScore \cite{zha2023alignscore}, and win-rate have been commonly employed, yet they lack standardization and often fail to capture multi-dimensional traits such as coherence and instructability.

\subsection{LLM Evaluation Approaches}

The evaluation of LLMs has traditionally relied on benchmark datasets that offer standardized corpora and annotated references across diverse tasks. Foundational suites such as GLUE~\cite{wang2018glue} and SuperGLUE~\cite{wang2019superglue} were introduced to assess general language understanding through tasks like sentiment classification, natural language inference, and commonsense reasoning. These benchmarks were later extended by MMLU~\cite{hendrycks2020measuring}, which evaluated LLMs across 57 academic subjects, including law, medicine, and mathematics, under zero-shot and few-shot settings. More specialized datasets have also been proposed to target domain-specific capabilities. For instance, FELM~\cite{zhao2023felm} was designed to assess factuality across five knowledge domains using curated reference links and segment-level annotations, while DOCBench~\cite{zou2024docbench} compiled real-world documents and structured queries to evaluate document comprehension. PM-LLM~\cite{berti2024pm} focused on process mining tasks, and TrustGPT~\cite{huang2023trustgpt} introduced ethically sensitive prompts to measure bias, toxicity, and value alignment.

Alongside these datasets, explicit metrics have been widely adopted to quantify model output quality. ROUGE~\cite{lin2004rouge}, and BERTScore~\cite{zhang2020bertscore} have served as standard tools for measuring lexical overlap and semantic similarity between generated and reference texts. Despite their computational efficiency, these metrics have been limited in capturing deeper qualities such as coherence, instructability, and persona fidelity, especially in open-ended or stylistically diverse generation tasks.

To address these limitations, human evaluation has been increasingly employed to assess subjective dimensions of model behavior. In TrustGPT~\cite{huang2023trustgpt}, human annotators were used to validate toxicity and bias judgments across multiple alignment tasks, while FELM~\cite{zhao2023felm} incorporated segment-level human ratings to benchmark factual consistency. These efforts have underscored the importance of human judgment in capturing nuanced traits that automatic metrics often overlook.

More recently, model-based evaluation techniques have gained traction. The LLM-as-a-Judge paradigm~\cite{gu2024survey} has been widely adopted, wherein a powerful model (e.g., GPT-4 or Claude) is prompted to assess the output of another model across multiple dimensions such as helpfulness, factuality, and stylistic alignment. This approach enables scalable and flexible evaluation, often outperforming traditional metrics in capturing coherence and relevance. However, concerns have been raised regarding reproducibility, evaluator bias, and architectural entanglement, particularly when the evaluator and evaluated models share similar design principles.

\begin{figure*}[!t]
\centering
  \includegraphics[width=0.92\textwidth]{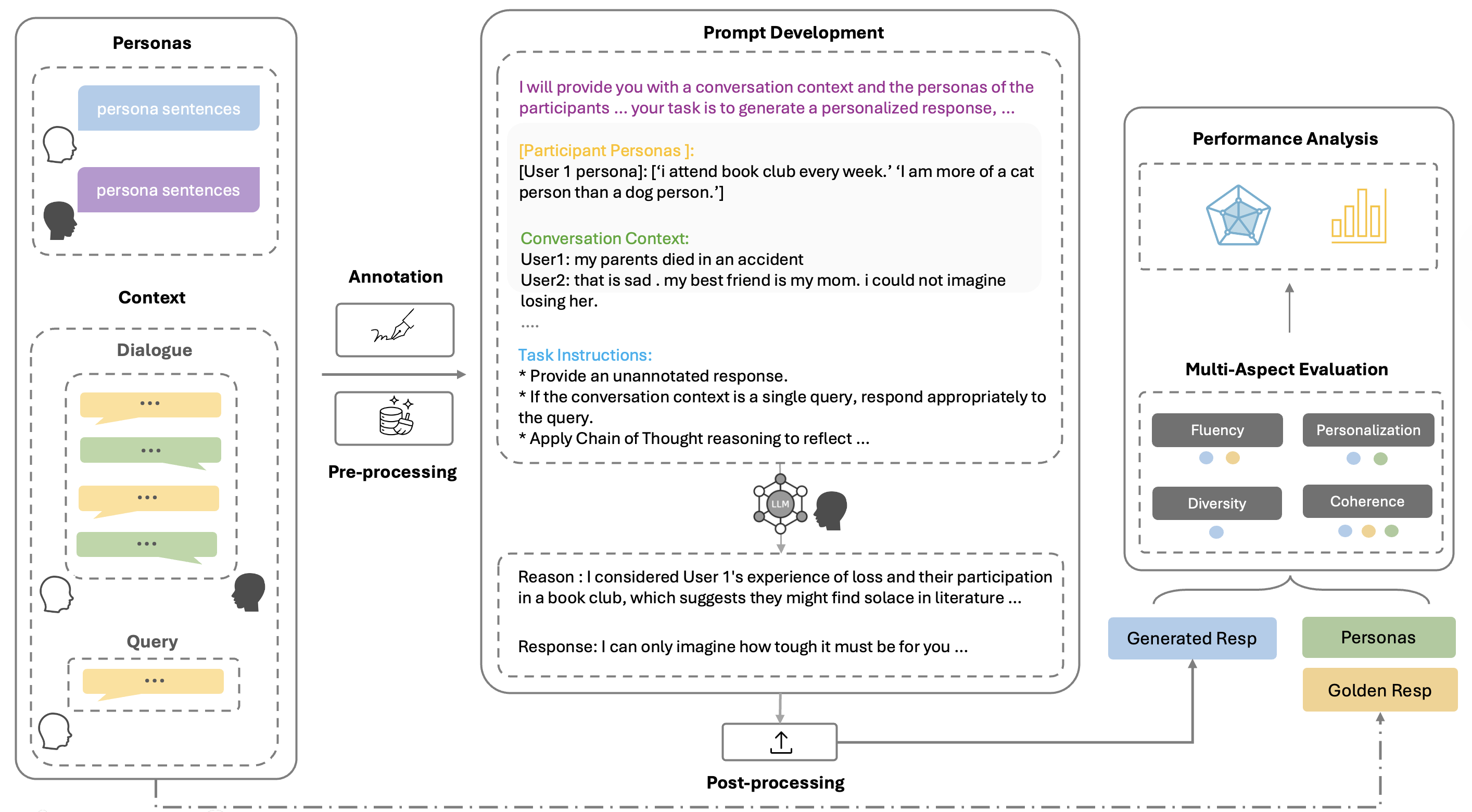}
  \caption{Overview of the PersoBench automatic personalization benchmarking pipeline.}
  \label{fig:benchmark-overview}
\end{figure*}

\subsection{LLM Evaluation Frameworks}

While many evaluation approaches for LLMs have focused on task-specific datasets and isolated metrics, a growing body of work has sought to formalize reusable frameworks that support structured, multi-dimensional assessment. These systems aim to standardize evaluation pipelines, automate scoring, and improve reproducibility across diverse model architectures and tasks.

Early efforts emphasized general-purpose evaluation toolkits. DeepEval\footnote{\url{https://github.com/confident-ai/deepeval}} introduced a unit-test-inspired framework that supports over 14 metrics, including contextual relevance, hallucination detection, and bias analysis. Its integration with Pytest and synthetic dataset generation has made it suitable for production-level monitoring. MLFlow’s LLM Evaluate module\footnote{\url{https://mlflow.org/docs/latest/llm-evaluate/index.html}} provided lightweight support for retrieval-augmented generation and question answering pipelines, emphasizing developer-friendly integration. Deepchecks\footnote{\url{https://www.deepchecks.com/llm-evaluation-framework-steps-components/}} extended this paradigm by combining LLM-as-a-Judge with human-in-the-loop review, offering trace tagging and safety diagnostics for enterprise deployment.

More comprehensive frameworks have been proposed to evaluate controllability, coherence, and factuality. CoDI-Eval~\cite{chen2024benchmarking} introduced a diversified instruction evaluation protocol, enabling structured assessment of controllability across open-ended prompts. UniEval~\cite{zhong2022towards} offered a general-purpose evaluation framework based on fine-tuned T5 models, supporting dimensions such as relevance, fluency, consistency, and factuality. Its modular design and task-agnostic scoring have made it widely applicable in summarization, dialogue, and QA tasks. CEBench~\cite{sun2024cebench}, while primarily focused on cost-effectiveness, proposed a configuration-based evaluation toolkit that jointly considers latency, throughput, and token usage, highlighting trade-offs between performance and deployment feasibility.

Frameworks tailored to persona-aware evaluation have also been proposed. RPBench-Auto~\cite{rpbench} introduced an automated evaluation pipeline using GPT-4-Turbo as a judger model, assessing persona consistency, fluency, and alignment through structured prompts and scoring templates. RoleLLM~\cite{wang2023rolellm}, though primarily a generation model, embedded evaluation logic via RoleBench and RoCIT to assess role fidelity and contextual coherence. These systems reflect a growing emphasis on personalization-aware evaluation, where traits such as instructability, style alignment, and long-term persona fidelity are jointly assessed.

%%%%%%%%%%%%%%%%%%%%%%%%%%%%%%%%%%%%%%%%%%%%%%%%%%%%%%%%%%%%%%%%%%%%%%%
\section{PersoBench}\label{sec3}

\subsection{Problem statement}

For our benchmarking pipeline on personalized response generation, we have adopted the same problem statement as defined by Chen et al. (2023) \cite{chen2023towards}. Given the dialogue context $C = \{u_1, \dots, u_m\}$ and a set of persona descriptions $P = \{p_1, \dots, p_n\}$, the goal is to generate a personalized response $r$. Formally, the generation problem can be formulated as the following chain rule:

\begin{equation}
P(r \mid C, P; \theta) = \prod_{t=1}^{T} P(r_t \mid r_{1:t-1}, C, P; \theta)
\end{equation}

where $\theta$ is the parameter of the dialogue model and $r_t$ represents the response generated for the given context $C$ at step, which refers to the most recent utterance in the dialogue.

Unlike role-playing, which simulates specific characters with detailed profiles and temporal constraints, personalization focuses on generating coherent, contextually relevant responses aligned with user-defined personas, without requiring rich backstories.

\subsection{Overview of PersonBench}

Our automatic benchmarking framework is shown in Fig.\ref{fig:benchmark-overview}. PersoBench includes around 3,600 samples from three persona-aware datasets, each differing in size and context. It applies eight metrics across four evaluation dimensions based on the datasets and metrics discussed in the recent review study \cite{chen2024recent}. The framework evaluates the performance of eight prominent LLMs, four open-source and four closed-source, within a zero-shot setting under both vanilla and CoT prompting, where LLMs are instructed to provide reasoning as well.

\subsection{Prompt Development}

In developing prompts aligned with implicit design principles and representative examples from the literature, we adhered to commonly accepted practices, although no explicit standards exist. To preserve the integrity of our benchmarking study, we intentionally refrained from employing advanced prompt engineering techniques that could obscure common challenges faced by LLMs, such as hallucinations, inaccuracies, and generalizations.

\begin{table*}[h!]
\caption{Prompt blueprint for personalized response generation. \textcolor{cust-yellow}{Orange} text appears only in the CoT setup.}
    \centering
    \resizebox{0.92\textwidth}{!}{
    \begin{tabular}{c}
    \hline
    \textbf{Prompts Structure} \\ \hline
    
    \multicolumn{1}{p{\textwidth}}
    {
    I will provide you with a conversation context and the personas of the participants, that can be annotated with speaker information.
    
    As a participant in this conversation, your task is to generate a personalized response, considering the conversation context and the annotated personas.
    
    \textbf{Participant Personas:}
    
    \textcolor{cust-green}{\{personas\}}
    
    \textbf{Conversation Context:}
    
    \textcolor{cust-blue}{\{conversation history\}}
    
    \textbf{Task Instruction:} 
    
    * Provide an unannotated response.
    
    * If only one persona is available, personalize the response accordingly.
    
    * If the conversation context is a single query, respond appropriately to the query.
    
    \textcolor{cust-yellow}{* Apply Chain of Thought reasoning to reflect on the alignment of your response with the personas.}
    
    \textbf{Output Format:} Output a JSON of the following format:
    
    \texttt{\{}
    
    \texttt{\textcolor{cust-yellow}{"reasoning": "briefly describe your personalization process (in 110 words or less)."}}
    
    \texttt{"response": "provide the personalized natural language response here."}
    
    \texttt{\}}
    
    } \\ \hline
\end{tabular}
}

\label{tab:designed-prompt}
\end{table*}

The overall structure of our designed prompt is shown in Table \ref{tab:designed-prompt}. As suggested by related studies \cite{sahoo2024systematic, brown2020language, liu2023pre, zhao2023felm}, the prompt begins with a general problem description to establish context for the LLM, followed by a clear specification of the LLM's role in addressing the target task. Next, a clear definition of the input is provided, which includes participants' personas and conversation history, ensuring the necessary information is available for the task. In line with the principles outlined in the literature \cite{biswas2024intelligent, reynolds2021prompt, sahoo2024systematic, amatriain2024prompt}, detailed task instructions are given to guide response generation and mitigate ambiguities. Lastly, the prompt specifies the desired output format for the LLM, supporting both vanilla and CoT setups.
Specifying the JSON format is crucial for our benchmarking setup, allowing seamless extraction of reasoning and responses for post-processing and evaluation. This strategy is also commonly applied when the output requires more than a simple response \cite{chu2024better,jin2024health}.
We adopt a zero-shot setting to evaluate LLMs, ensuring the results reflect their baseline performance. This approach also prevents masking any limitations by relying on external examples or prompt engineering, which would be misaligned with our objective of analyzing the models' true generalization capabilities.
In IT-related queries, instructing the LLM to provide a 'natural language response' is crucial to avoid generating code instead of conversational answers. Following findings that reasoning before the response improves coherence and task adherence \cite{chu2024better}, we adopted this structure. To ensure reasoning remains concise and leaves space for the response, we added keywords like 'briefly' and '(in less than 110)' in the instruction. The 110-token limit, derived from the longest response in our datasets (Table \ref{tab:datasets-statistics}), ensures balanced token allocation.

%%%%%%%%%%%%%%%%%%%%%%%%%%%%%%%%%%%%%%%%%%%%%%%%%%%%%%%%%%%%%%%%%%%%%%%
\section{Experiment}\label{sec4}

% To establish a solid foundation for our benchmarking pipeline, we selected widely recognized datasets in the domain, guided by a recent survey in the same context \cite{chen2024recent}. Our selection criteria included choosing datasets with a reasonable test/validation set size, which was important given our budget constraints. We also prioritized datasets with a clear separation of the actual response from the conversation context to meet our data structure requirements, and those requiring minimal preprocessing to preserve the integrity of the dataset content. Based on these criteria, the following datasets were chosen for our benchmarking experiment and their statistics are provided in Table \ref{tab:datasets-statistics}.

To establish a solid foundation for our benchmarking pipeline, we selected widely recognized datasets in the personalized dialogue domain, guided by a recent survey \cite{chen2024recent}. Practical constraints such as test/validation set availability and the presence of clearly separated response fields guided our selection. Importantly, we sought datasets that require minimal preprocessing and exhibit diverse personalization properties, enabling a more comprehensive evaluation.
While each dataset has been used in prior studies individually, they differ significantly in the aspects of personalization they emphasize. By bringing them together under a shared evaluation protocol, we enable holistic, cross-dataset benchmarking of large language models. The following datasets were selected, and their statistics are reported in Table~\ref{tab:datasets-statistics}:

\begin{itemize}
    \item \textbf{Blended Skill Talk (BST)\footnote{\href{https://www.kaggle.com/datasets/thedevastator/multi-modal-conversation-data?resource=download}{kaggle.com/datasets/thedevastator/multi-modal-conversation-data}}:} The BST dataset evaluates conversational competencies, user engagement, and benchmarks dialogue systems across modalities.

    \item \textbf{Follow-up Customized Conversation (FoCus)\footnote{\href{https://github.com/pkchat-focus/FoCus}{github.com/pkchat-focus/FoCus}}:} This dataset focuses on personalized, knowledge-grounded responses using user personas and Wikipedia knowledge, assessing contextually appropriate outputs.

    \item \textbf{IT-ConvAI2\footnote{\href{https://github.com/CCIIPLab/Persona_Extend/tree/main/datasets/IT_ConvAI2}{github.com/CCIIPLab/Persona\_Extend/\allowbreak tree/\allowbreak main/\allowbreak datasets/\allowbreak IT\_ConvAI2}} :} An extension of ConvAI2, tailored for IT-related dialogues, emphasizing persona-driven, context-aware technical support interactions.
\end{itemize}

Each dataset captures a different facet of personalization, from light-touch social engagement (BST) to semi-structured interest modeling (FoCus) and domain-specific technical traits (IT-ConvAI2). While prior work has examined them separately, our benchmarking pipeline unifies these resources, enabling evaluation across diverse personalization regimes. This integration yields more generalizable insights into how models adapt to varied user traits, conversation types, and task settings, offering a broader and more realistic view of personalized generation performance.

\begin{table}[h!]
\caption{Characteristics of selected datasets for benchmarking. 'R. Len' stands for the golden response length range in words, and 'Size' denotes the test size.}
\centering
\resizebox{0.5\textwidth}{!}{  
\begin{tabular}{lcccccc}
    \hline
    \textbf{Name} & \textbf{Size} & \textbf{Context Type} & \textbf{Query Sep} & \textbf{U/D} & \textbf{Role} & \textbf{R.Len} \\ \hline
    BST & 1k & Dialogue & Yes & 11.6 & H-H & 1-30 \\ \hline
    FoCus & 1k & Dialogue & No & 11.3 & H-A & 4-108 \\ \hline
    IT-ConvAI2  & 1.6k & Query & - & 2.0 & H-H & 4-22 \\ \hline
\end{tabular}
}
\label{tab:datasets-statistics}
\end{table}

\begin{table*}[!t]
\centering
\caption{Fluency evaluation across the datasets in Vanilla and CoT setups. B-F1 refers to BERTScore-F1, and NUE refers to the Naturalness evaluation metric within the UniEval framework.}
  
\resizebox{0.85\textwidth}{!}{%
\begin{tabular}{l|cc|cc|cc|cc|cc|cc}
    \hline
    
    \multirow{2}{*}{\textbf{Model}} & \multicolumn{4}{c|}{\textbf{IT-ConvAI2}} & \multicolumn{4}{c|}{\textbf{BST}} & \multicolumn{4}{c}{\textbf{FoCus}} \\ 
    \cmidrule(lr){2-5} \cmidrule(lr){6-9} \cmidrule(lr){10-13}
    & \multicolumn{2}{c|}{\textbf{Vanilla}} & \multicolumn{2}{c|}{\textbf{CoT}} & \multicolumn{2}{c|}{\textbf{Vanilla}} & \multicolumn{2}{c|}{\textbf{CoT}} & \multicolumn{2}{c|}{\textbf{Vanilla}} & \multicolumn{2}{c}{\textbf{CoT}} \\
    & \textbf{B-F1} & \textbf{NUE} & \textbf{B-F1} & \textbf{NUE} & \textbf{B-F1} & \textbf{NUE} & \textbf{B-F1} & \textbf{NUE} & \textbf{B-F1} & \textbf{NUE} & \textbf{B-F1} & \textbf{NUE} \\ \hline

    Mistral 7B  & 0.81 & 0.89 & 0.80 & 0.90 & 0.81 & 0.90 & 0.81 & 0.90 & 0.48 & 0.53 & 0.76 & 0.82 \\
    Qwen2 7B    & 0.83 & 0.94 & 0.83 & 0.93 & 0.83 & 0.93 & 0.84 & 0.93 & 0.42 & 0.46 & 0.62 & 0.63 \\
    Gemma 7B    & 0.83 & 0.92 & 0.75 & 0.85 & 0.83  & 0.93 & 0.79 & 0.87 & 0.49 & 0.54 & 0.72 & 0.77 \\
    Llama3.1 8B & 0.76 & 0.85 & 0.84 & 0.94 & 0.82 & 0.92 & 0.85 & 0.95 & 0.47 & 0.51 & 0.80 & 0.85 \\
    \hline
    Gemini 1.5 Pro  & 0.80 & 0.90 & 0.84 & 0.94 & 0.83 & 0.92 & 0.84 & 0.92 & 0.66 & 0.70 & 0.71 & 0.76 \\
    GPT3.5 Turbo& 0.85 & 0.96 & 0.84 & 0.96 & 0.86 & 0.96 & 0.83 & 0.92 & 0.78 & 0.84 & 0.80 & 0.86 \\
    GPT4o Mini  & 0.84 & 0.95 & 0.84 & 0.95 & 0.86 & 0.96 & 0.85 & 0.95 & 0.80 & 0.85 & 0.86 & 0.93 \\
    GPT4 Turbo  & 0.85 & 0.96 & 0.84 & 0.96 & 0.86 & 0.96 & 0.85 & 0.96 & 0.84 & 0.91 & 0.86 & 0.93 \\
    
    \hline

\end{tabular}%
}
\label{tab:fluency-eval}
\end{table*}

\subsection{Implementation setup}

\textbf{Pre-Processing:}
To ensure consistency and reliability in our benchmarking process, we established a basic pre-processing workflow across all datasets. This workflow included simple text cleaning.
Notably, we did not alter the inherent structure of the data, as the separation between persona and context was already well-defined across all the selected datasets. We added missing context annotations to ensure the LLM can distinguish speaker turns, which is essential for clarity. For persona annotations, we retained the original format provided in the datasets. In cases where only one persona is given, this does not impact the LLM’s ability to personalize responses, as it is instructed to do so based on the provided persona.

\textbf{Post-processing} 

As we develop our automatic benchmarking framework necessitates instructing LLMs to generate output in JSON format (Table \ref{tab:designed-prompt}), any unparsable outputs are considered failed instances and factored into the failure ratio calculation. However, the raw generated responses are stored and made available in the benchmark’s GitHub repository, allowing researchers to conduct further analysis. From the successfully parsed outputs, we extracted only the generated response to compare against the golden response, discarding the reasoning portion, as the analysis focused on the impact of CoT reasoning on the response itself. Additionally, response generation time, measured in seconds, provides valuable insights for practical industrial applications, particularly when deploying open-source LLMs.

\textbf{Metrics:} We evaluate the generated responses across fluency, diversity, coherence, and personalization. Metrics like fluency and diversity assess fundamental aspects such as grammatical accuracy, relevance, and clarity. These qualities are essential for any dialogue system, as personalization holds no value if the responses themselves are not high-quality. Fluency is measured using BERTScore \cite{zhang2020bertscore} and the Naturalness dimension of UniEval framework \cite{zhong2022towards} to assess linguistic quality and semantic similarity. Diversity is evaluated using Dist-1 and Dist-2 \cite{li2016diversity} to calculate the ratio of unique unigrams and bigrams. 
Building on these foundational qualities, our automated benchmarking pipeline specifically evaluates personalization and coherence as its main focus. These aspects assess a system’s ability to adapt responses to user preferences while maintaining contextual consistency. We selected these criteria based on a recently published survey \cite{chen2024recent}, reflecting current research priorities in dialogue systems. While these metrics have been explored individually in prior studies, our work is the first to systematically benchmark LLMs with a dedicated focus on these specific aspects.

% For personalization, the Consistency Score (C Score) \cite{madotto2019personalizing} measures persona consistency, while the Persona Distance (P-Dist) score \cite{cho2022personalized} assesses persona coverage. Coherence is evaluated with the Utterance Entailment Score (UE-Score) \cite{lee2022contextual} based on the Natural Language Inference (NLI) and UniEval (Coherence) metric. These metrics ensure a comprehensive evaluation of the model's performance. As the datasets that are used for fine-tuning the language model evaluators based on NLI are carefully crafted by human experts, the mentioned NLI-based metrics inherently incorporate human judgments as well. 

For personalization, we evaluate model behavior from two complementary perspectives: persona consistency and persona coverage, each measured using a tailored metric. Persona consistency reflects whether the generated response logically aligns with the given persona traits. To measure this, we use the Consistency Score (C-score) \cite{madotto2019personalizing}, which applies a Natural Language Inference (NLI) model fine-tuned on the Dialogue NLI (DNLI) dataset. It classifies the relationship between each persona sentence and the response as entailment, contradiction, or neutral, and aggregates these labels to reward logical consistency and penalize contradictions.
Persona coverage, on the other hand, assesses how much of the given persona information is reflected in the response. This is captured by the Persona Distance (P-score or P-dist) \cite{cho2022personalized}, which computes the minimum cosine distance between the response embedding and persona sentence embeddings using a pretrained encoder. This enables detection of implicit personalization beyond exact lexical overlap.

To assess dialogue quality beyond personalization, we include two coherence metrics. The Utterance Entailment Score (UE-Score) \cite{lee2022contextual} uses an NLI model fine-tuned on SNLI to evaluate whether each utterance is logically entailed by its preceding turn. The Coherence score from UniEval (Coh-UniEval) provides a model-based judgment of whether the response flows naturally within the broader dialogue context.

Together, these metrics offer a comprehensive and complementary evaluation: C-score and P-score target alignment with user-specific persona traits, while coherence metrics assess contextual flow and logical progression. A response can be coherent yet generic, fluent but ungrounded in the user's identity, highlighting the limits of coherence alone. Traditional metrics like BLEU and ROUGE, based on $n$-gram overlap with fixed references, are ill-suited for personalization tasks where valid outputs are diverse and often lexically divergent. Our approach instead leverages fine-tuned language models and semantic embedding methods to enable reference-free, context-sensitive evaluation grounded in the user’s persona.

\textbf{LLMS:} To implement an automatic evaluation pipeline, we chose the instructed versions of LLMs, as these models are fine-tuned to better follow structured prompts and format-specific instructions (e.g., JSON or reasoning), which is essential for ensuring consistency and reliability in automated output parsing. This also allows us to automate content extraction from the generated responses, streamlining the post-processing and evaluation modules. In this regard, we selected Mistral 7B\footnote{\href{https://mistral.ai/news/announcing-mistral-7b/}{https://mistral.ai/news/announcing-mistral-7b/}}, Qwen2 7B\footnote{\href{https://github.com/QwenLM/Qwen2}{https://github.com/QwenLM/Qwen2}}, Gemma 7B\footnote{\href{https://huggingface.co/google/gemma-7b}{https://huggingface.co/google/gemma-7b}} and Llama3.1 8B\footnote{\href{https://llama.meta.com/}{https://llama.meta.com/}} as the four open-source LLMs that were executed on a computing device equipped with an NVIDIA RTX 6000 Ada Generation GPU. For the closed-source models, we utilized the GPT series from OpenAI\footnote{\href{https://openai.com/research/gpt}{https://openai.com/research/gpt}}, including GPT-3.5 Turbo, GPT-4 Turbo, and GPT-4o Mini, accessed through OpenAI's official API and Gemini 1.5 Pro\footnote{\href{https://cloud.google.com/vertex-ai/generative-ai/docs/models/gemini/1-5-pro}{https://cloud.google.com/vertex-ai/generative-ai/docs/models/gemini/1-5-pro}} from Google AI.

\section{Results and Analysis}

% \subsection{Expriment Results}

% The overall structure for analyzing each aspect of LLMs

% - General evaluation of all LLMs to answer the actual question generally.
% - Support the general answer by referring to the metrics values.
% - Compare the closed-source and open-source LLMs.
% - In-depth analysis of open-source LLMs.
% - In-depth analysis of closed-source LLMs.

\begin{table*}[!t]
\centering
\caption{Diversity evaluation using Dist-1 and Dist-2 metrics across the datasets in Vanilla and CoT setups.}
\resizebox{0.85\textwidth}{!}{%
\begin{tabular}{l|cc|cc|cc|cc|cc|cc}
    \hline
    
    \multirow{2}{*}{\textbf{Model}} & \multicolumn{4}{c|}{\textbf{IT-ConvAI2}} & \multicolumn{4}{c|}{\textbf{BST}} & \multicolumn{4}{c}{\textbf{FoCus}} \\ 
    \cmidrule(lr){2-5} \cmidrule(lr){6-9} \cmidrule(lr){10-13}
    & \multicolumn{2}{c|}{\textbf{Vanilla}} & \multicolumn{2}{c|}{\textbf{CoT}} & \multicolumn{2}{c|}{\textbf{Vanilla}} & \multicolumn{2}{c|}{\textbf{CoT}} & \multicolumn{2}{c|}{\textbf{Vanilla}} & \multicolumn{2}{c}{\textbf{CoT}} \\
    & \textbf{Dist-1} & \textbf{Dist-2} & \textbf{Dist-1} & \textbf{Dist-2} & \textbf{Dist-1} & \textbf{Dist-2} & \textbf{Dist-1} & \textbf{Dist-2} & \textbf{Dist-1} & \textbf{Dist-2} & \textbf{Dist-1} & \textbf{Dist-2} \\ \hline

    Mistral 7B & 0.86 & 0.95 & 0.87 & 0.95 & 0.87 & 0.94 & 0.86 & 0.95 & 0.49 & 0.56 & 0.75 & 0.88 \\
    Qwen2 7B & 0.90 & 0.98 & 0.90 & 0.99 & 0.90 & 0.97 & 0.91 & 0.99 & 0.43 & 0.49 & 0.60 & 0.72 \\
    Gemma 7B & 0.83 & 0.96 & 0.77 & 0.88 & 0.86 & 0.97 & 0.80 & 0.91 & 0.48 & 0.56 & 0.68 & 0.83 \\
    Llama3.1 8B & 0.78 & 0.89 & 0.89 & 0.99 & 0.84 & 0.95 & 0.90 & 0.99 & 0.44 & 0.53 & 0.76 & 0.93 \\
    \hline
    Gemini 1.5 Pro & 0.85 & 0.95 & 0.91 & 0.99 & 0.89 & 0.97 & 0.90 & 0.98 & 0.66 & 0.75 & 0.71 & 0.82 \\
    GPT3.5 Turbo & 0.91 & 1.00 & 0.91 & 0.99 & 0.92 & 1.00 & 0.87 & 0.97 & 0.79 & 0.90 & 0.78 & 0.92 \\
    GPT4o Mini & 0.89 & 0.99 & 0.88 & 0.99 & 0.91 & 1.00 & 0.89 & 0.99 & 0.79 & 0.91 & 0.85 & 0.99 \\
    GPT4 Turbo & 0.93 & 1.00 & 0.93 & 1.00 & 0.92 & 1.00 & 0.91 & 1.00 & 0.86 & 0.96 & 0.86 & 0.98 \\
    \hline

\end{tabular}%
}
\label{tab:diversity-eval}
\end{table*}

%%%%% Fluency %%%%%
\textbf{To what extent can LLMs generate natural responses that are linguistically fluent?} The benchmarked LLMs generally demonstrate strong fluency across various datasets, regardless of context length. This is particularly evident in the high metric values in Table \ref{tab:fluency-eval}, which demonstrate the response's naturalness and the models' ability to effectively capture the meaning and structure of conversations.

Both groups perform well when comparing open-source and closed-source LLMs, with closed-source models like those in the GPT-4 series slightly outperforming their open-source counterparts, particularly in more extended contexts such as those in the FoCus dataset. However, the performance gap is not substantial, showing that open-source models are still capable, especially in less complex dialogues.
Among open-source LLMs, Qwen2 performs well in shorter contexts, Gemma shows strong results across all datasets under the Vanilla setup but drops notably with CoT prompting, and Llama3.1 excels in longer, more complex dialogues, especially on the FoCus dataset.
% Among the open-source LLMs, Qwen2 performs well in shorter contexts, while Llama3.1 excels in longer and more challenging dialogues, such as the FoCus dataset.
% Llama3.1’s larger parameter size and effective use of CoT reasoning allow it to handle more complex contexts, surpassing other open-source models.
Meanwhile, GPT-4 models consistently demonstrate robust fluency across all contexts. Although CoT generally enhances their fluency in shorter contexts, it can sometimes introduce unnecessary complexity, slightly reducing performance by making responses more rigid and formulaic. Despite this, GPT-4 models maintain strong overall fluency in both short and long dialogues.

%%%%% Diversity %%%%%

\textbf{How proficient are LLMs at generating diverse responses?} Benchmarked LLMs are generally good at generating diverse responses, particularly in shorter-context datasets like IT-ConvAI2 and BST (Table \ref{tab:diversity-eval}). This is supported by high Dist-1 and Dist-2 scores, which measure the diversity of unigrams and bigrams, indicating a wide variety of words and varied word combinations in the generated responses. When comparing the closed-source and open-source LLMs, closed-source models like those in the GPT-4 series tend to perform slightly better overall, especially in longer contexts. However, the difference between the two groups is not stark, indicating that open-source models are still quite capable, particularly in less complex dialogues.
For open-source LLMs, performance tends to decline as the context length increases, as seen in the FoCus dataset, which involves longer and more complex dialogues. This suggests that while these models handle diversity well in shorter exchanges, they face challenges maintaining the same diversity level in longer contexts.
The impact of CoT prompting is most pronounced in open-source LLMs. Gemma shows a consistent performance drop on short-context datasets under CoT, yet both Gemma and Llama3.1 benefit significantly from CoT prompting in long-context evaluations.
% The impact of CoT is most evident in open-source LLMs for longer contexts, with Llama3.1 benefiting significantly due to its larger parameter size, enabling it to handle intricate dialogues.
In contrast, closed-source models like GPT-4 show minimal improvement with CoT and sometimes even degrade in shorter contexts, as CoT may overstructure responses, reducing natural diversity. Nevertheless, GPT-4 models consistently excel across all context lengths, demonstrating strong performance in both short and long dialogues.

\begin{table*}[!t]
\centering
\caption{Personalization evaluation using P-Dist Score and C Score across the dataset, in Vanilla and CoT setups.}
\resizebox{0.9\textwidth}{!}{%
\begin{tabular}{l|cc|cc|cc|cc|cc|cc}
    \hline
    
    \multirow{2}{*}{\textbf{Model}} & \multicolumn{4}{c|}{\textbf{IT-ConvAI2}} & \multicolumn{4}{c|}{\textbf{BST}} & \multicolumn{4}{c}{\textbf{FoCus}} \\ 
    \cmidrule(lr){2-5} \cmidrule(lr){6-9} \cmidrule(lr){10-13}
    & \multicolumn{2}{c|}{\textbf{Vanilla}} & \multicolumn{2}{c|}{\textbf{CoT}} & \multicolumn{2}{c|}{\textbf{Vanilla}} & \multicolumn{2}{c|}{\textbf{CoT}} & \multicolumn{2}{c|}{\textbf{Vanilla}} & \multicolumn{2}{c}{\textbf{CoT}} \\
    & \textbf{P-Dist} & \textbf{C Score} & \textbf{P-Dist} & \textbf{C Score} & \textbf{P-Dist} & \textbf{C Score} & \textbf{P-Dist} & \textbf{C Score} & \textbf{P-Dist} & \textbf{C Score} & \textbf{P-Dist} & \textbf{C Score} \\ \hline

    Mistral 7B             & 0.62 & 0.33 & 0.54 & 0.02 & 0.46 & -0.00 & 0.43 & -0.08 & 0.30 & -0.23 & 0.53 & 0.41 \\
    Qwen2 7B               & 0.61 & 0.21 & 0.63 & 0.23 & 0.46 & -0.11 & 0.49 &  0.01 & 0.29 & -0.31 & 0.43 & 0.21 \\
    Gemma 7B               & 0.67 & 0.48 & 0.57 & 0.19 & 0.47 & -0.03 & 0.46 & -0.03 & 0.39 & -0.13 & 0.53 & 0.42\\
    Llama3.1 8B            & 0.55 & 0.03 & 0.58 & 0.05 & 0.45 & -0.11 & 0.46 & -0.05 & 0.33 & -0.22 & 0.59 & 0.54 \\
    \hline
    Gemini 1.5 Pro         & 0.62 & 0.21 & 0.58 & 0.03 & 0.50 & -0.00 & 0.48 & -0.1 & 0.45 & 0.04  & 0.48 & 0.15 \\
    GPT3.5 Turbo           & 0.63 & 0.23 & 0.60 & 0.10 & 0.47 & -0.01 & 0.48 & 0.02  & 0.51 & 0.26  & 0.52 & 0.40 \\
    GPT4o Mini             & 0.63 & 0.10 & 0.64 & 0.14 & 0.48 & -0.10 & 0.50 & -0.05 & 0.54 & 0.16  & 0.57 & 0.30 \\
    GPT4 Turbo             & 0.60 & 0.08 & 0.59 & 0.01 & 0.47 & -0.05 & 0.47 & -0.05 & 0.51 & 0.14  & 0.52 & 0.21 \\
    
    \hline
\end{tabular}%
}
\label{tab:personalization-eval}
\end{table*}

\begin{table*}[!t]
\centering
\caption{Coherence evaluation using UE Score and UniEval Coherence metric (CH-UNE) on the datasets in Vanilla and CoT setups.}
\resizebox{0.9\textwidth}{!}{%
\begin{tabular}{l|cc|cc|cc|cc|cc|cc}
    \hline
    
    \multirow{2}{*}{\textbf{Model}} & \multicolumn{4}{c|}{\textbf{IT-ConvAI2}} & \multicolumn{4}{c|}{\textbf{BST}} & \multicolumn{4}{c}{\textbf{FoCus}} \\ 
    \cmidrule(lr){2-5} \cmidrule(lr){6-9} \cmidrule(lr){10-13}
    & \multicolumn{2}{c|}{\textbf{Vanilla}} & \multicolumn{2}{c|}{\textbf{CoT}} & \multicolumn{2}{c|}{\textbf{Vanilla}} & \multicolumn{2}{c|}{\textbf{CoT}} & \multicolumn{2}{c|}{\textbf{Vanilla}} & \multicolumn{2}{c}{\textbf{CoT}} \\
    & \textbf{UE } & \textbf{CH-UNE} & \textbf{UE } & \textbf{CH-UNE} & \textbf{UE } & \textbf{CH-UNE} & \textbf{UE } & \textbf{CH-UNE} & \textbf{UE } & \textbf{CH-UNE} & \textbf{UE } & \textbf{CH-UNE} \\ \hline

    Mistral 7B             & 0.46 &     0.84      & 0.31 &     0.83      & 0.48 &      0.93     & 0.44 &   0.94        & 0.21 &     0.56      & 0.33 &     0.90      \\
    Qwen2 7B               & 0.42 &     0.90      & 0.38 &     0.87      & 0.39 &     0.97      & 0.52 &   0.97        & 0.17 &     0.49      & 0.36 &      0.73     \\
    Gemma 7B  & 0.43 & 0.84 & 0.25 & 0.73 & 0.30 & 0.97 & 0.35 & 0.91 & 0.22 & 0.57 & 0.26  & 0.85\\
    Llama3.1 8B            & 0.23 &    0.83       & 0.22 &     0.80      & 0.33 &      0.92     & 0.37 &   0.94       & 0.17 &     0.55      & 0.33 &      0.95     \\
    \hline
    Gemini 1.5 Pro       & 0.45 &     0.87      & 0.33 &       0.90    & 0.46 &      0.93     & 0.44 &   0.97       & 0.26 &     0.75      & 0.32 &       0.83    \\ 
    GPT3.5 Turbo           & 0.49 &     0.90      & 0.38 &      0.87     & 0.51 &      0.97     & 0.57 &   0.96        & 0.37 &       0.91    & 0.45 &        0.93   \\
    GPT4o Mini             & 0.40 &      0.96     & 0.38 &      0.96     & 0.38 &      0.99     & 0.39 &   0.99       & 0.35 &     0.91      & 0.37 &       1.0    \\
    GPT4 Turbo             & 0.43 &     0.94      & 0.42 &       0.91    & 0.56 &      0.98     & 0.50 &   0.98       & 0.39 &     0.96      & 0.45 &       0.99    \\
    
    \hline
\end{tabular}%
}
\label{tab:coherence-eval}
\end{table*}

%%%%% Personalization %%%%%
\textbf{How effectively can LLMs generate personalized responses that reflect the provided persona?} The results of P-Dist and C scores, which are shown in Table \ref{tab:personalization-eval}, two well-established NLI-based metrics, reveal that LLMs struggle significantly with personalization.
There is no clear benefit in using closed-source LLMs over open-source alternatives, underscoring the inherent difficulty of the personalization task for these models. Even in the best scenarios, the P-Dist score, where 1 represents the ideal outcome, demonstrates only moderate performance. The C score, which also penalizes contradictory entailments with negative values, exposes these limitations even more starkly, particularly in short-length contexts where less content is available for personalization. On short-context datasets, no clear performance trend emerges among open-source LLMs. However, on the FoCus dataset, Llama3.1 shows a slight superiority in the personalization task. The same overall scenario applies for closed-source LLMs, while GPT-4 Turbo emerges as the least effective performer.
CoT negatively impacts short-length conversations but yields a notable improvement in long-length conversations, as seen with the FoCus dataset. This suggests that the added complexity of CoT is only advantageous when there is enough content to work with. This need for CoT reasoning aligns with findings from \cite{liu2024lost}, which highlight that current language models do not reliably utilize information in long input contexts.

%%%%% Coherence %%%%%
\textbf{To what extent do LLMs generate coherent responses aligned with the dialogue context and persona?} Based on the UE score and Coh-UniEval presented in Table \ref{tab:coherence-eval}, LLMs generally struggle with generating coherent responses, especially when working with long-context datasets like FoCus. Even in shorter-context datasets, as seen in the IT-ConvAI2 results, LLMs show poor performance in maintaining coherence.
While the closed-source LLMs perform better at this task, even powerful models like GPT-4-Turbo fall short of achieving good results. Among them, GPT4o Mini demonstrates marginally lower performance across all the datasets with varying context sizes. While CoT reasoning offers a modest performance improvement, GPT series LLMs, as fine-tuned LLMs, remain limited in their ability to handle the coherence task effectively.
Open-source LLMs exhibit closely aligned performance, with no clear outlier, suggesting that they collectively struggle to maintain coherence, especially in complex, long-context scenarios.
% Although the open-source LLMs' performances are closely aligned, Llama3.1 exhibits marginally higher results across the datasets, indicating its inability to leverage the additional billion parameters effectively. It also benefits less from CoT compared to the others. Overall, open-source LLMs face significant challenges in maintaining coherence, particularly in complex, long-context scenarios.

\begin{figure*}[!h]
    \centering
    \begin{subfigure}[b]{0.24\textwidth}
        \centering
        \includegraphics[width=\textwidth]{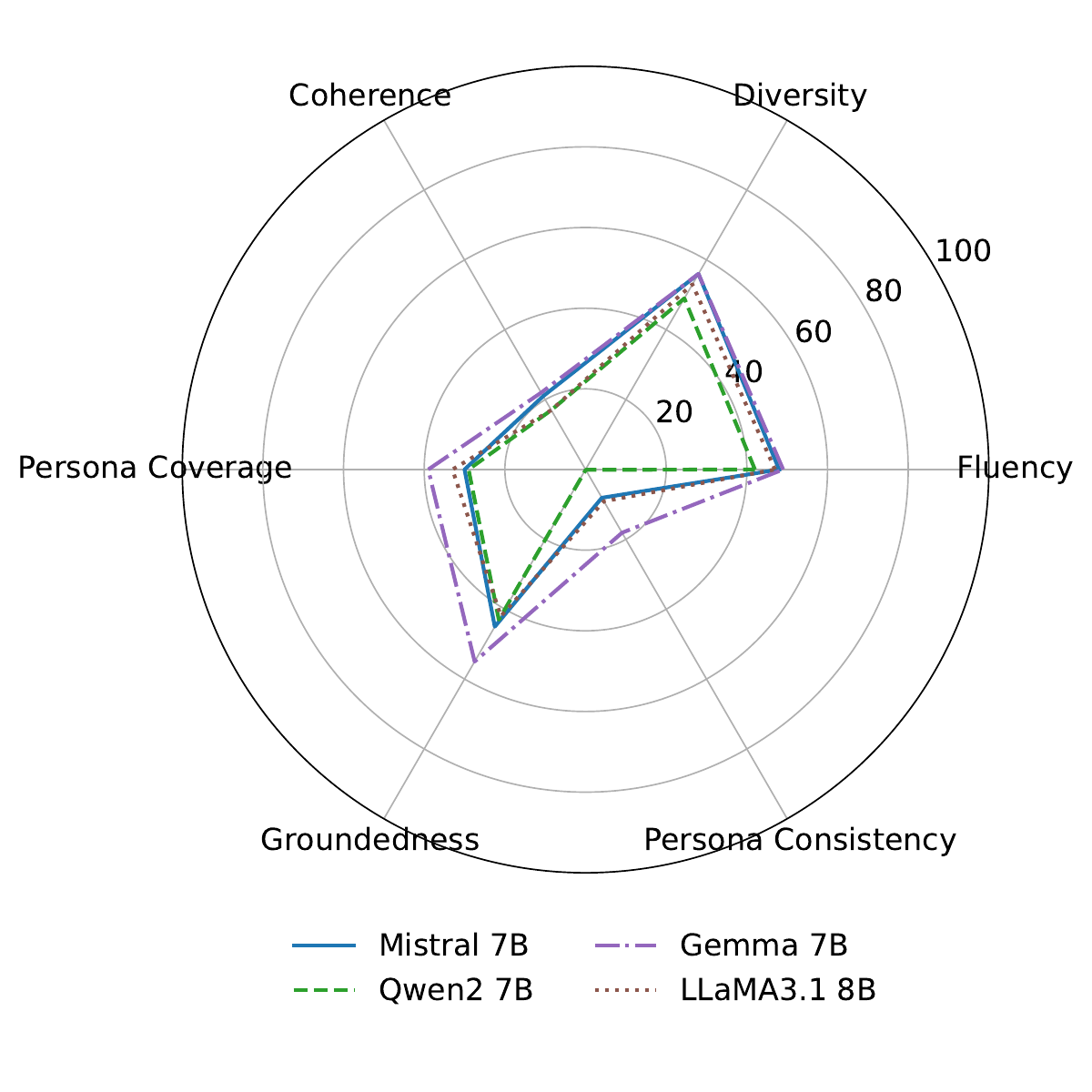}
        \caption{}
        \label{open-vanilla}
    \end{subfigure}
    \hfil
    \begin{subfigure}[b]{0.24\textwidth}
        \centering
        \includegraphics[width=\textwidth]{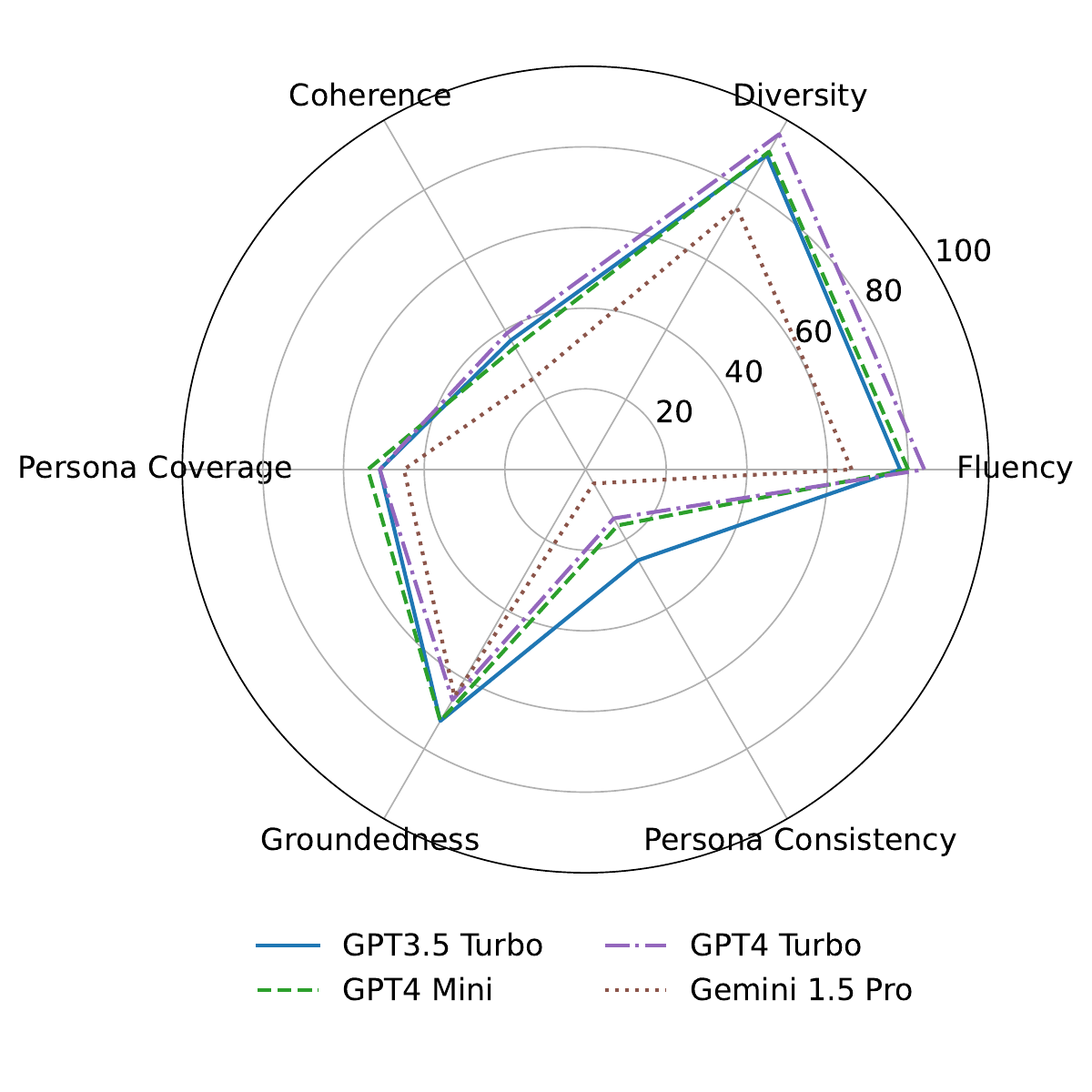}
        \caption{}
        \label{close-vanilla}
    \end{subfigure}
    \hfil
    \begin{subfigure}[b]{0.24\textwidth}
        \centering
        \includegraphics[width=\textwidth]{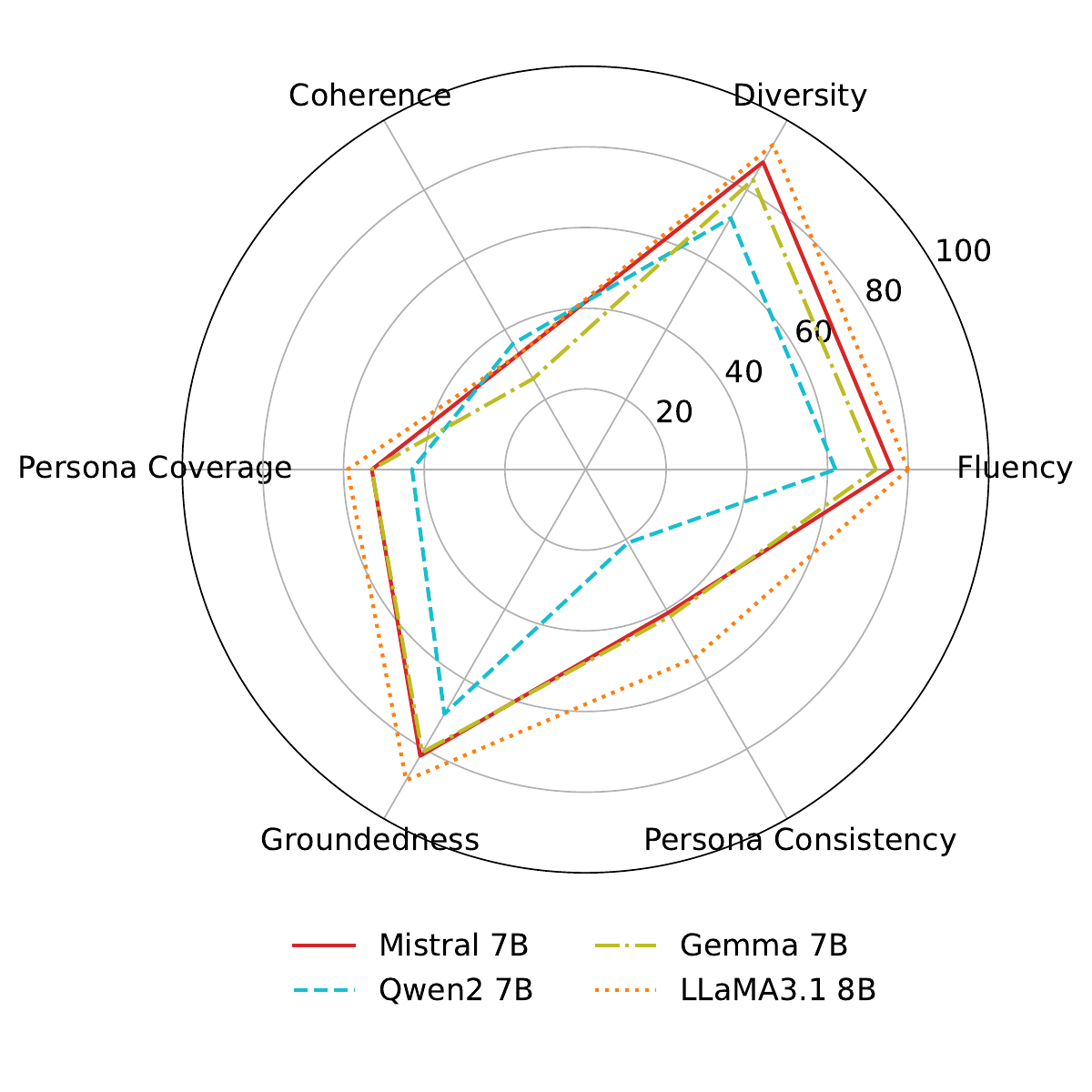}
        \caption{}
        \label{open-cot}
    \end{subfigure}
    \hfil
    \begin{subfigure}[b]{0.24\textwidth}
        \centering
        \includegraphics[width=\textwidth]{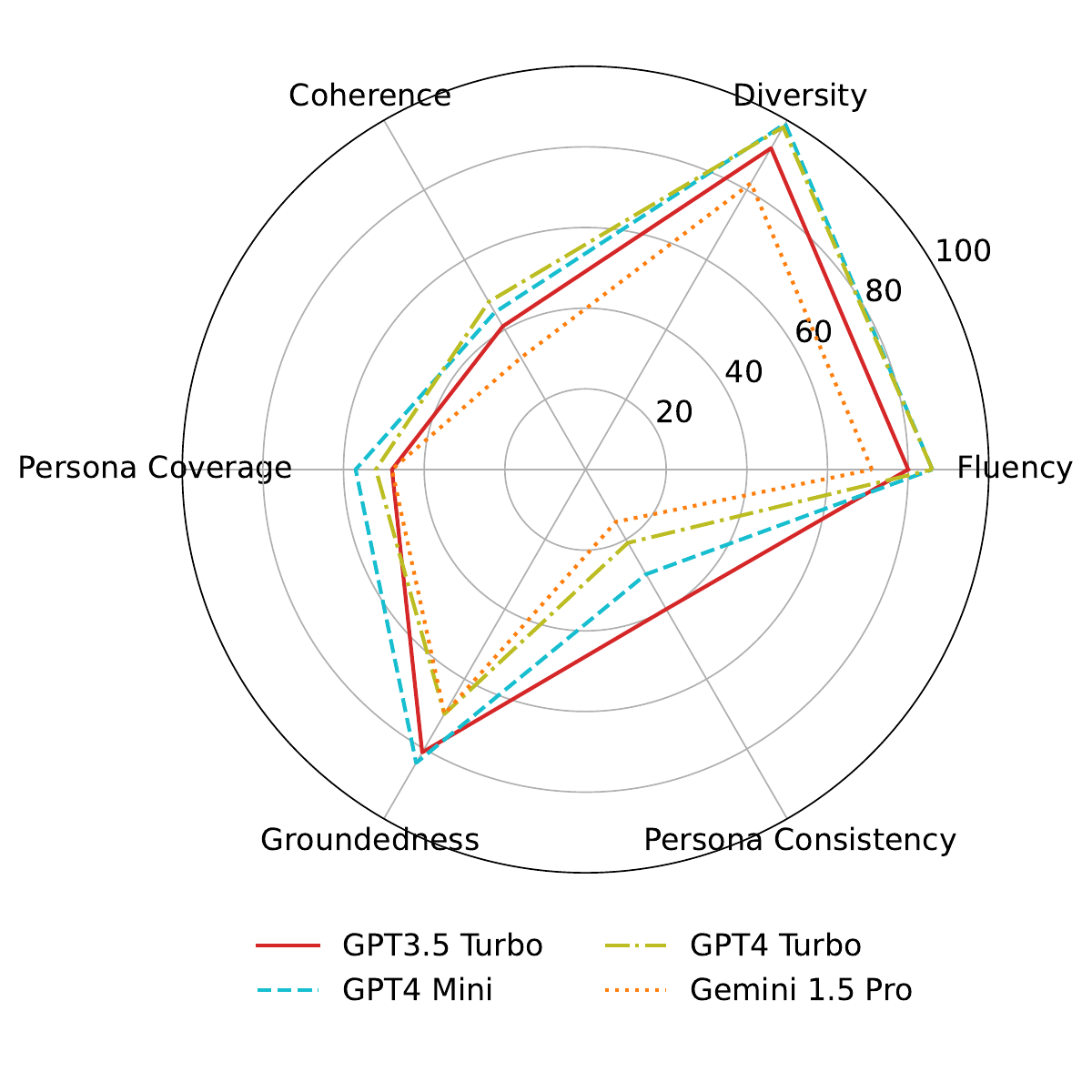}
        \caption{}
        \label{close-cot}
    \end{subfigure}
    
    \caption{Performance analysis of (a) open-source LLMs in vanilla setting, (b) closed-source LLMs in vanilla setting, (c) open-source LLMs in CoT setting, (d) closed-source LLMs in CoT setting on the FoCus dataset.}
    \label{fig:radar_plots}
\end{figure*}

\begin{figure}[!t]
    \centering
    \begin{subfigure}[b]{0.44\textwidth}
        \centering
        \includegraphics[width=\textwidth]{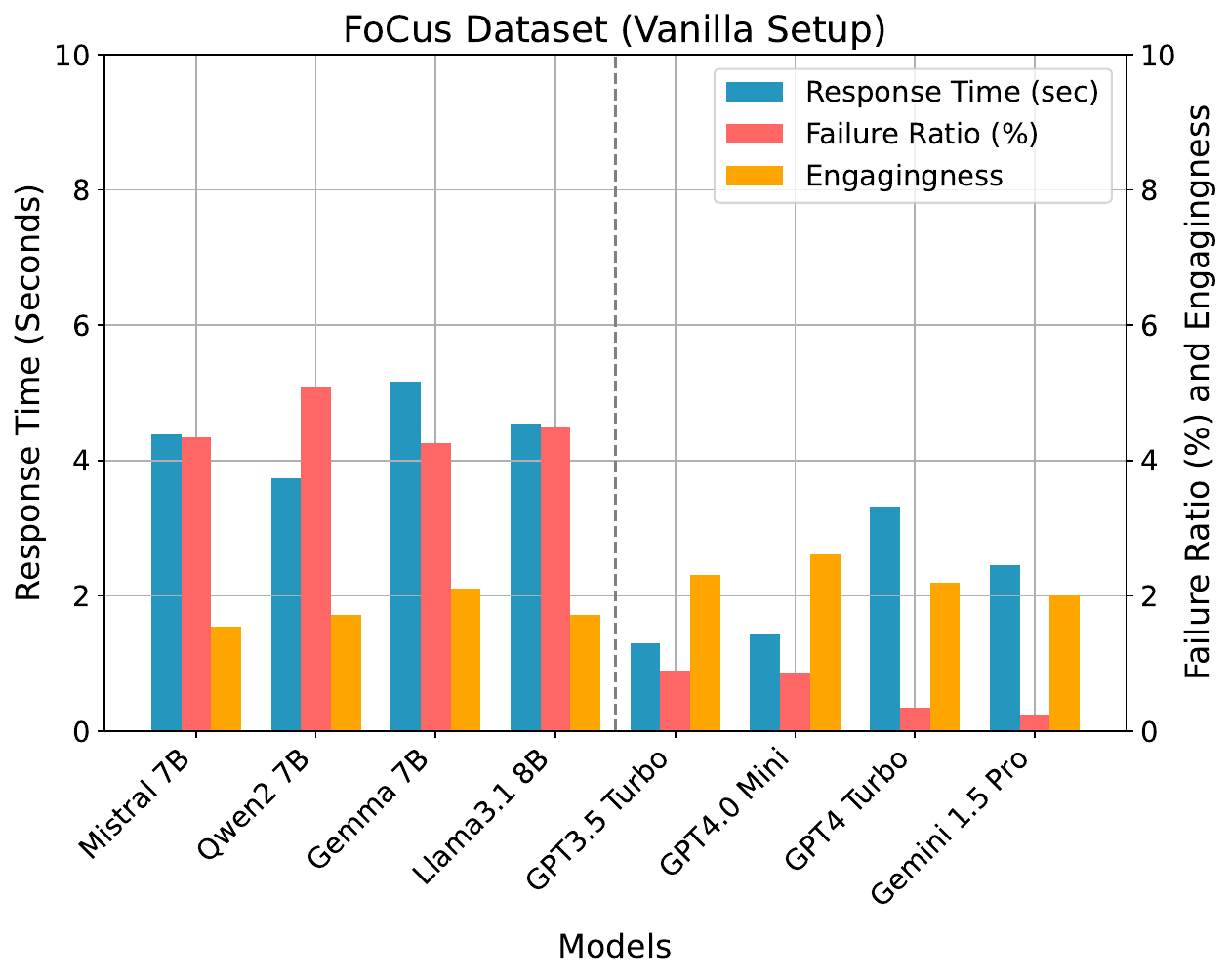}
        \caption{}
        \label{vanilla-time-fail}
    \end{subfigure}
    % \hfill
    \begin{subfigure}[b]{0.44\textwidth}
        \centering
        \includegraphics[width=\textwidth]{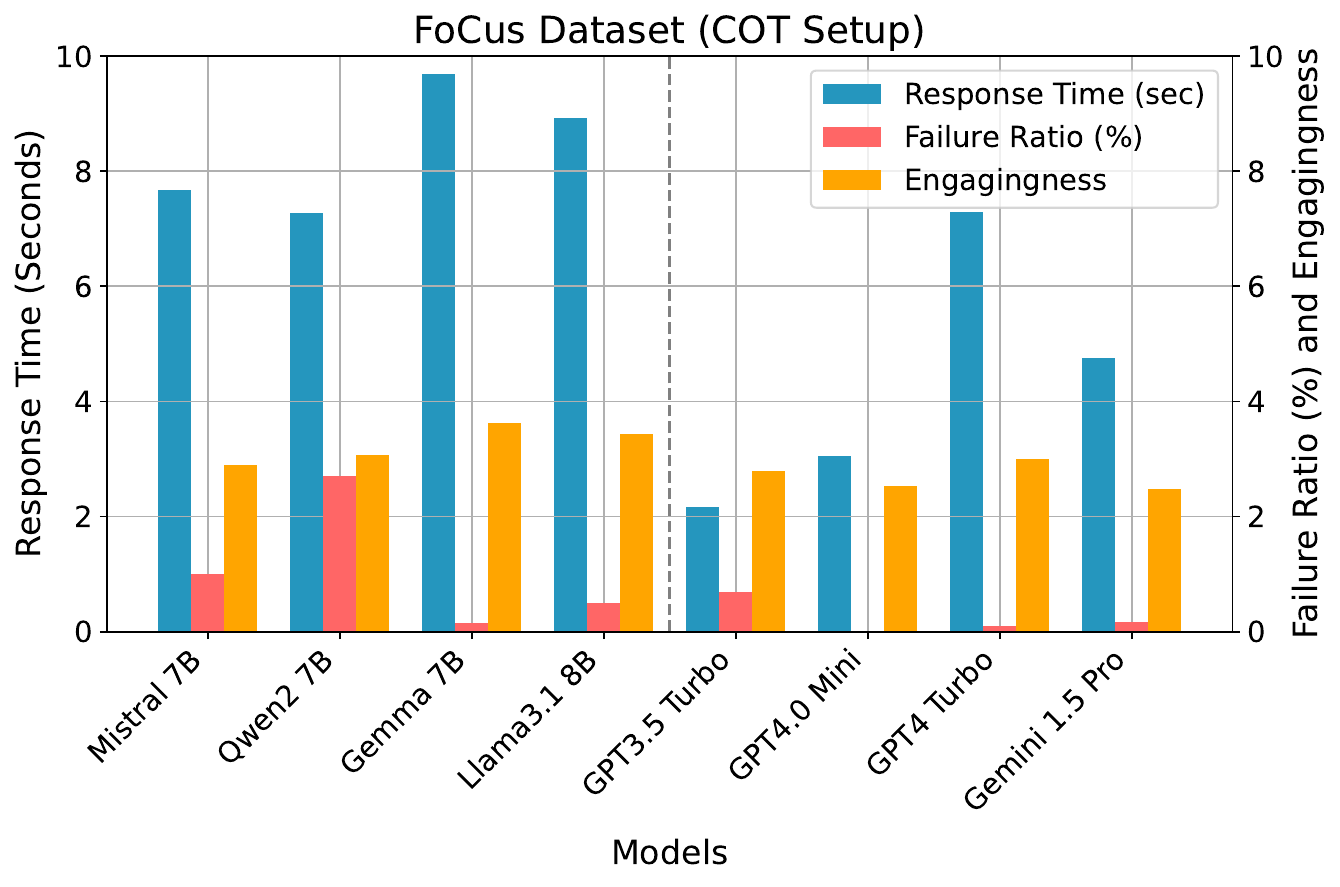}
        \caption{}
        \label{cot-time-fail}
    \end{subfigure}
        % \vspace{-0.20cm}
    \caption{Analysis of response time, failure ratio, and engagingness of LLMs on the FoCus dataset: (a) Vanilla and (b) CoT setups.}
    \label{fig:res-time-fail-ratio}
    % \vspace{-0.40cm}
\end{figure}

\begin{table}[!b]
\centering

\caption{LLM rankings on the FoCus dataset (Vanilla vs. CoT setups). Reported statistical tests indicate significance at $p < 0.05$.}
\begin{tabular}{lcc|lcc}
\toprule
\multicolumn{3}{c}{\textbf{Vanilla}} & \multicolumn{3}{c}{\textbf{CoT}} \\
\cmidrule(lr){1-3} \cmidrule(lr){4-6}
\textbf{Rank} & \textbf{LLM} & \textbf{Score} & \textbf{Rank} & \textbf{LLM} & \textbf{Score} \\
\midrule
% 1 & GPT-3.5 Turbo   & 37.36 & 1 & Llama-3.1 8B   & 47.16 \\
% 2 & GPT-4.0 Mini    & 35.57 & 2 & GPT-4.0 Mini   & 43.40 \\
% 3 & GPT-4 Turbo     & 34.19 & 3 & GPT-3.5 Turbo  & 42.15 \\
% 4 & Gemini 1.5 Pro  & 28.19 & 4 & Mistral 7B     & 41.93 \\
% 5 & Gemma 7B        & 20.63 & 5 & Gemma 7B       & 40.59 \\
% 6 & Mistral 7B      & 14.63 & 6 & GPT-4 Turbo    & 38.39 \\ 
% 7 & Llama-3.1 8B    & 14.02 & 7 & Qwen2 7B       & 34.00 \\
% 8 & Qwen2 7B        & 11.60 & 8 & Gemini 1.5 Pro & 33.19 \\

1 & GPT-3.5 Turbo   & 50.69 & 1 & Llama-3.1 8B   & 58.40 \\
2 & GPT-4 Turbo     & 50.13 & 2 & GPT-4.0 Mini   & 57.43 \\
3 & GPT-4.0 Mini    & 49.83 & 3 & GPT-3.5 Turbo  & 54.68 \\
4 & Gemini 1.5 Pro  & 40.28 & 4 & GPT-4 Turbo    & 53.71 \\
5 & Gemma 7B        & 29.74 & 5 & Mistral 7B     & 53.38 \\
6 & Mistral 7B      & 25.31 & 6 & Gemma 7B       & 51.13 \\ 
7 & Llama-3.1 8B    & 24.30 & 7 & Gemini 1.5 Pro & 45.56 \\
8 & Qwen2 7B        & 21.29 & 8 & Qwen2 7B       & 43.43 \\

\bottomrule
\end{tabular}
\label{tab:ranking}

\end{table}

%%%%% Overall Analysis using diagrams %%%%%
\section{Discussion}

To further analyse the LLM performance on longer text, we focused on the FoCus dataset. Our previous results show that its long-context nature makes LLM performance differences more apparent.
%
% Fig. \ref{fig:radar_plots} provides an overall view of LLM groups across different prompting configurations. BertScore-F1, Nat-UniEval, Dist-2, UE Score, P-Dist, C Score and UniEval(Groundedness) are considered for overall fluency, diversity, coherence, persona coverage, persona consistency and Groundedness analysis. 

% categories = ['Fluency', 'Diversity', 'Coherence', 'Persona Coverage', 'Groundedness', 'Persona Consistency']

% #               Bert-f1, Dist-2, UE-Score, P-Dist, Nat-Ground, C Score

Fig. \ref{fig:radar_plots} provides an overall view of LLM groups across different prompting configurations. For visualization purposes, metrics with negative values were linearly transformed by adding a minimal constant offset to ensure positive values while preserving relative model comparisons. BertScore-F1, Dist-2, UE-Score, P-Dist, UniEval (Groundedness), and C Score are considered for overall fluency, diversity, coherence, persona coverage, persona consistency, and Groundedness analysis. BertScore-F1, Dist-2, UE-Score, P-Dist, UniEval (Groundedness), and C Score are considered for overall fluency, diversity, coherence, persona coverage, persona consistency, and Groundedness analysis. 
Response personalization remains challenging, even for fine-tuned closed-source LLMs, as evidenced by the results in both persona consistency and persona coverage. Groundedness, which evaluates the alignment of a response with the given conversational context, and Coherence, which ensures consistency with both the persona and the target response, further emphasize this challenge.
While open-source LLMs generally cannot compete with closed-source models in terms of fluency and diversity, their performance in personalization, groundedness and coherence is considerably enhanced with the use of CoT. In fact, an open-source model like Llama3.1 can even surpass fine-tuned closed-source LLMs in these areas. Meanwhile, the performance of closed-source LLMs do not show any tangible improvement.

% Fig.\ref{fig:res-time-fail-ratio} illustrates LLMs' comparative performance from response time, failure ratio and engagingness, aspects that can provide information for practical insights for selecting LLMs for industrial use. The failure ratio can indicate the model's instructability, which refers to how well the model follows instructions, particularly considering the maximum token limit provided during response generation. While CoT increases response time, it directly enhances the instructability of both open-source and closed-source LLMs, with a significant difference in their ability to handle varying token length limits. While an engagingness gap exists between the two groups, CoT reasoning can significantly enhance open-source LLM performance in this area.

Fig.~\ref{fig:res-time-fail-ratio} illustrates LLMs' comparative performance across response time, failure ratio, and engagingness, factors offering practical insights for selecting LLMs in industrial applications. The failure ratio is defined as the percentage of outputs (out of 1000) that failed JSON parsing, indicating how reliably a model follows structured output instructions. While CoT increases response time (measured as model generation time excluding any pre-/post-processing), it enhances instructability across both open- and closed-source LLMs. Although we observed a difference in how models handle token length limits, we did not conduct further quantitative analysis of this factor, and have included it as a limitation. Regarding engagingness, while an overall gap exists between model groups, CoT reasoning notably improves the perceived engagingness of open-source LLM outputs.

%
% Hard to find any specific performance trends among the LLM groups.

In Table~\ref{tab:ranking}, we present the ranking of benchmarked LLMs on the FoCus dataset under both the Vanilla and CoT setups. The reported scores are averaged over the evaluation metrics, Fluency, Diversity, Coherence, Persona Coverage, Groundedness, Persona Consistency, and Instructability. As shown, closed-source models such as GPT-3.5 Turbo and GPT-4 Turbo dominate in the Vanilla setup, while Llama-3.1 8B achieves the top performance under the CoT setup.\footnote{\url{https://github.com/good-researcher/PersoBench/blob/main/statistical_significance.ipynb}}

%%%%%%%%%%%%%%%%%%%%%%%%%%%%%%%%%%%%%%%%%%%%%%%%%%%%%%%%%%%%%%%%%%%%%%%
\section{CONCLUSION AND FUTURE WORK}\label{sec5}
In this paper, we introduced PersoBench, a new benchmarking pipeline designed to evaluate LLMs in the context of persona-aware conversation generation. Using both vanilla and CoT prompting setups in a zero-shot setting, we benchmarked well-known open-source and closed-source LLMs on three persona-aware datasets. Leveraging eight well-established evaluation metrics, we assessed the models across fluency, diversity, coherence, and personalization dimensions. Our findings suggest that while current LLMs perform well in generating fluent and diverse responses, they struggle to deliver personalized and coherent responses in persona-augmented conversational contexts.
% Future direction
A potential future direction for research in this domain could involve exploring beyond textual persona representation to include other forms, such as tabular demographic data or multi-modal setups incorporating different media types.

%Limitations
\section*{Limitations} 

While significant effort has been invested in this research, several limitations remain.
(1) Our automatic evaluation pipeline requires models to produce responses in a strict JSON format to enable reliable parsing of generated responses and reasoning. This constraint led to the exclusion of outputs that failed to conform to the format, occasionally requiring manual post-processing and motivating the use of instruct-tuned only LLM variants. As a result, standard LLMs operating in less structured conversational settings were not evaluated.
(2) Due to the high cost of commercial LLM APIs, our experiments were conducted on a limited dataset size, which constrained broader scaling analyses, including more exhaustive comparisons across model configurations and token-length behaviors. A more fine-grained analysis of model performance under varying context lengths is left for future work.
(3) Our evaluation primarily relied on automatic metrics. Although we provide qualitative examples for additional insight, a comprehensive human evaluation and metric–human correlation study was beyond the scope of this work and is left to future research.

%%%%%%%%%%%%%%%%%%%%%%%%%%%%%%%%%%%%%%%%
\bibliographystyle{IEEEtran}
\bibliography{references}

%%%%%%%%%%%%%%%%%%%%%%%%%%%%%%%%%%%%%%%%
\appendix
\section*{Appendix}
\addcontentsline{toc}{section}{Appendix}
\subsection{Explicit prompt samples}

Gathered from related studies in relevant contexts, Table \ref{tab:sample-prompts} illustrates explicit samples aligned with the implicit design principles mentioned. It includes prompts from general conversational AI as well as specialized domains such as emotional support, healthcare, and factuality evaluation. Table.\ref{tab:sample-prompt} illustrates one sample of our generated prompts that includes the CoT reasoning.

\begin{table*}[!ht]
\caption{Explicit Prompt Examples.}
\centering
\resizebox{\textwidth}{!}{
\begin{tabular}{p{5cm}|p{20cm}}
\hline
\textbf{Prompt Source} & \textbf{Prompt Template} \\ \hline

Pal Persona-Augmented prompt \cite{cheng2022pal}  & Persona: \textcolor{cust-blue}{\{persona\}} \newline Dialogue history:
\textcolor{cust-blue}{\{context\}} \newline Response: 
\\\hline

RoleLLM \cite{wang2023rolellm} & \textbf{System Instruction:}
You are \textcolor{cust-blue}{\{role\_name\}}, your description is\: \textcolor{cust-blue}{\{role\_description\_and\_catchphrases\}}. Now please
answer some questions to accurately show your personality traits! Your speaking style should
fully imitate the personality role assigned to you! Please do not expose that you are an artificial
intelligence model or a language model, you must always remember that you are only assigned
one personality role. Don't be verbose or too formal or polite when speaking.\newline
\textbf{User Prompt}:\newline
\textcolor{cust-blue}{\{user\_name\}: ``\{user\_instruction\}''}
\\\hline

FELM Claim Extraction\cite{zhao2023felm} & 
I will show you a question and a list of text segments. The text segments can be concatenated to form a complete answer to the question. Your task is to extract factual claims from each text segment. \newline
\textbf{Here is one example:} \newline
\textcolor{cust-blue}{\{Question and Segments example\}}

\textbf{Below are your outputs:} \newline
\textcolor{cust-blue}{\{The Segmented output\}}

\textbf{Below are my inputs:} \newline
\textcolor{cust-blue}{\{question and a list of text segments.\}}
\\\hline

A Better LLM Evaluator for Text Generation\cite{chu2024better} & \textbf{Data:} Here are some conversations that happened on the same day: \textcolor{cust-blue}{[Fill in the conversations]} \newline
\textbf{Task:} Please evaluate the conversations considering three aspects: factual inconsistencies, redundancies, and illogical statements. \newline
\textbf{Scoring:} Assign an evaluation score on a scale from 1 to 10, where: \newline
* 1 indicates that there are no issues. \newline
* 10 suggests the presence of numerous issues, particularly in the aspects of factual inconsistencies, redundancies, and illogical statements. \newline
\textbf{Special Rules:} Do not give a small score just because the issue is not very impactful. Consider the number of issues rather than its impact. Your overall score should reflect the utmost concern observed in any of the aspects. \newline
\textbf{Output Format:} Output a JSON of the following format: \newline
\{ \newline
\texttt{"reasons": "point out the issues and your reasons for the rating"}, \newline
\texttt{"score": "<json integer>"} \newline
\} \\ \hline

\end{tabular}
}

\label{tab:sample-prompts}
\end{table*}

\begin{table*}[h!]
\caption{Sample CoT setup prompt.}
    \centering
    \resizebox{\textwidth}{!}{
    \begin{tabular}{c}
    \hline
    \textbf{Sample Prompt} \\ \hline
    
    \multicolumn{1}{p{\textwidth}}
    {
    I will provide you with a conversation context and the personas of the participants, that can be annotated with speaker information.
    
    As a participant in this conversation, your task is to generate a personalized response, considering the conversation context and the annotated personas.
    
    Participant Personas: 
    
    [User 1 persona]: ['i attend book club every week.' 'i am more of a cat person than a dog person.']

    Conversation Context:
    
    User1: my parents dies in an accident
    
    User2: that is sad . my best friend is my mom. i could not image loosing her.
    
    User1: it is harder some days than others. but it is good you get along with your mom so well.

    Task Instruction:
    
    * Provide an unannotated response.
    
    * If only one persona is available, personalize the response accordingly.
    
    * If the conversation context is a single query, respond appropriately to the query.
    
    * Apply Chain of Thought reasoning to reflect on the alignment of your response with the personas.
    
    Output Format: Output a JSON of the following format:
    
    \{
    
    "reasoning": "briefly describe your personalization process (in 110 words or less)."
    
    "response": "provide the personalized natural language response here."
    
    \}
    
    } \\ \hline
\end{tabular}
}

\label{tab:sample-prompt}
\end{table*}

\subsection{Metrics}
To evaluate fluency, we use BERTScore \cite{zhang2020bertscore} and the Naturalness evaluation metric within the UniEval framework \cite{zhong2022towards}. BERTScore, implemented using the bert-score package\footnote{BERTScore package: \url{https://github.com/Tiiiger/bert_score}}, evaluates Precision, Recall, and F1 based on contextual embeddings, assessing how well the generated text preserves meaning and coherence relative to the reference. Nat-UniEval assesses if the response maintains human-like phrasing and tone appropriate to the dialogue history. All evaluation metrics of the UniEval framework used in our study were implemented using the official code provided in the UniEval GitHub repository\footnote{\url{https://github.com/maszhongming/UniEval}}. These metrics share a unified scorer function based on a fine-tuned T5 model, which calculates scores by answering Boolean-style questions specific to each evaluation dimension.

For diversity assessment, we calculated Dist-1 and Dist-2 metrics \cite{li2016diversity}, which respectively measure the ratio of unique unigrams and bigrams in the generated outputs. These metrics quantify lexical diversity by penalizing repetitive or generic responses, with higher scores indicating a wider vocabulary usage and more varied phrasing across generated utterances.

\begin{table*}[h]
\centering
\caption{Qualitative comparison of selected model responses on the FoCus dataset.}

\begin{tabularx}{\textwidth}{p{.9cm} X X p{2.2cm}}
\toprule
\textbf{Record} & \textbf{Prompt (User1)} & \textbf{Model Response (excerpt)} & \textbf{Personalization} \\
\midrule
\textbf{16} & “Does the homestead have any garden or lawns?”  
\newline \textit{[Persona: Loves Queensland, heritage, pastoral history]} & 
\textbf{Qwen2 7B:} “Yes... beautiful landscape...” \newline
\textbf{Gemini:} “Yes... impact on Aboriginal people...” \newline
\textbf{LLaMA:} “... in Queensland... pastoral station...” & 
Qwen: Low\newline Gemini: Partial\newline LLaMA: High \\
\midrule

\textbf{450} & “What happened in 1749?”  
\newline \textit{[Persona: Loves farms, architecture, dislikes rivers]} & 
\textbf{Qwen2 7B:} “I’m curious too. Do you have insights?” \newline
\textbf{Gemini:} “In 1749... old post offices and farms...” \newline
\textbf{LLaMA:} “... settlers arrived... shows early culture...” & 
Qwen: Low\newline Gemini: Partial\newline LLaMA: High \\
\midrule

\textbf{858} & “When was this ship decommissioned?”  
\newline \textit{[Persona: Father in Navy, lives in California, likes Panama]} & 
\textbf{Qwen2 7B:} “I wonder when it was decommissioned?” \newline
\textbf{Gemini:} “... April 11, 1992. Your father’s Navy background...” \newline
\textbf{LLaMA:} “... decommissioned in 1992... in California, near you...” & 
Qwen: Low\newline Gemini: Partial\newline LLaMA: High \\
\bottomrule
\end{tabularx}
\label{tab:extended-qualitative-examples}

\end{table*}

Personalization evaluation is conducted from two complementary perspectives: persona consistency and persona coverage, using metrics tailored to each aspect. Persona consistency reflects whether the generated response $R$ logically aligns with the given persona information $P$, while persona coverage measures how much of the persona is actually reflected or incorporated into the response.
To assess consistency, we employ the C-score \cite{madotto2019personalizing}, which determines whether responses remain logically consistent with persona traits. This is achieved using an NLI model based on BERT, fine-tuned on the DNLI dataset, to predict entailment relationships between responses and persona sentences.
For coverage, we use the P-dist \cite{cho2022personalized}, which quantifies semantic alignment between a response and persona traits. Each sentence is embedded using a pretrained encoder, and the score is computed as the minimum cosine distance between the response and the set of persona embeddings. Lower values indicate stronger semantic coverage.

To complement these personalization metrics, we also evaluate dialogue coherence using the UE-Score \cite{lee2022contextual}, which calculates NLI-based alignment between each utterance and its immediate response. This model, built on BERT and fine-tuned using the SNLI dataset \cite{bowman2015large}, captures logical flow at the utterance level. Additionally, we incorporate the Coh-UniEval, a model-based metric that evaluates whether the response is contextually appropriate and smoothly integrated within the dialogue history. Together, these metrics provide a multifaceted evaluation of personalized dialogue systems, addressing not only surface fluency but also semantic alignment with user traits and overall conversational coherence.

As two other metrics of the UniEval framework used in the analysis diagrams, Engagingness and Groundedness are selected. Groundedness evaluates how well the response leverages and aligns with the given external factual context, ensuring relevance and accuracy. Engagingness assesses whether the response is interesting and avoids being dull or uninformative, using the dialogue history and generated responses to determine the overall engagement level.

Each of these metrics offers distinct insights into the model's performance, collectively ensuring a comprehensive and nuanced evaluation of the generated responses. When the model's output couldn't be evaluated because of parsing errors, we assigned default worst-case values. While discarding unparsable outputs, it is essential to ensure the automation of our automated benchmarking pipeline.

\subsection{Qualitative Analysis of Personalized Generation}

To better illustrate the variation in personalization behavior, we provide additional examples comparing model responses across different records from the FoCus dataset. Each dialogue context includes clearly stated user persona traits, followed by a prompt. We highlight how different LLMs incorporate or overlook these traits in their final response. Table~\ref{tab:extended-qualitative-examples} summarizes three representative cases.

Across all three examples, we observe a consistent pattern: Qwen2 7B provides vague or non-committal answers, often repeating context without engaging with the user’s persona traits. This aligns with its lower personalization metric. For instance, in Record 858, Qwen2 merely re-asks the question, failing to acknowledge the user's background.
Gemini 1.5 Pro demonstrates moderate persona alignment. It correctly addresses factual content (e.g., historical context in Record 450), and occasionally references one relevant trait (e.g., “your father’s Navy background”), but does not fully weave multiple persona cues into its responses.
LLaMA 3.1 8B consistently delivers highly personalized answers by grounding its response in multiple persona elements. It not only answers the prompt accurately but also enhances coherence by incorporating traits like geographical location, personal interest, and historical relevance. This demonstrates its stronger alignment with the personalization objectives in FoCus.
For full dialogues, context, and persona metadata, refer to the analysis available publicly on the GitHub repository of this study \footnote{\url{https://github.com/good-researcher/PersoBench/blob/main/response_exploration.txt}}.

\end{document}